\documentclass{article}
\usepackage{spconf,amsmath,graphicx,multirow}
\usepackage{caption}

\usepackage{pifont}
\usepackage[dvipsnames]{xcolor}
\definecolor{limegreen}{HTML}{32CD32}
\newcommand{\cmark}{\textcolor{limegreen}{\ding{51}}}%
\newcommand{\xmark}{\textcolor{red}{\ding{55}}}%

\title{Optimizing Latent Space Directions For GAN-based Local Image Editing}
\name{Ehsan Pajouheshgar, Tong Zhang, and Sabine S\"usstrunk}
\address{School of Computer and Communication Sciences, EPFL, Switzerland}

\begin{document}
%
\twocolumn[{%
\renewcommand\twocolumn[1][]{#1}%
\maketitle
\begin{center}
    \centering
    \captionsetup{type=figure}
    \includegraphics[trim={63pt 225pt 300pt 50pt},clip,width=1.0\textwidth]{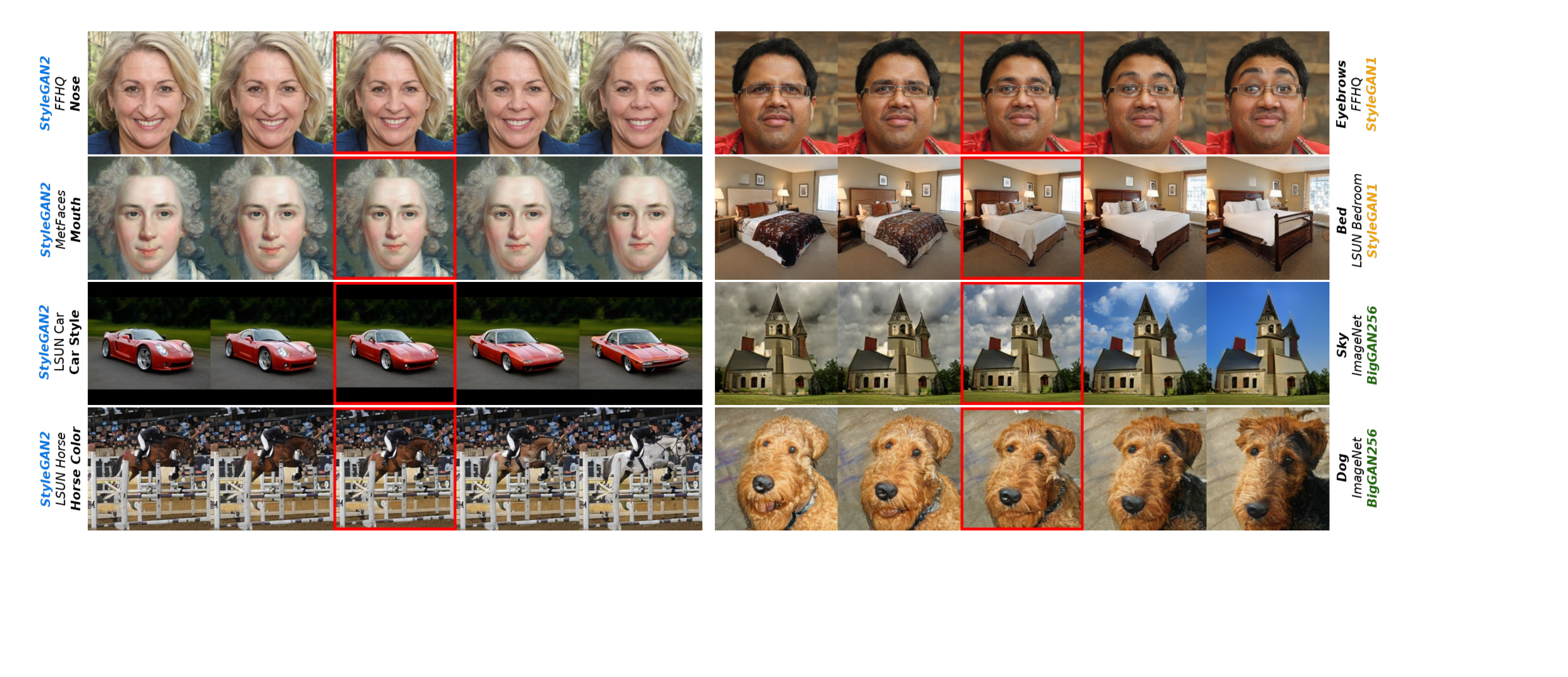}
    \vspace{-18pt}
    \captionof{figure}{Examples of localized edits performed by our method on GAN-generated images. 
    The images to the left and right of the middle one are results of moving the latent code in directions discovered by our method to edit each semantic part.
    }
    \label{fig:main} 

\end{center}%
}]
\begin{abstract}
Generative Adversarial Network (GAN) based localized image editing can suffer from ambiguity between semantic attributes. We thus present a novel objective function to evaluate the locality of an image edit. By introducing the supervision from a pre-trained segmentation network and optimizing the objective function, our framework, called Locally  Effective  Latent  Space  Direction (LELSD), is applicable to any dataset and GAN architecture. Our method is also computationally fast and exhibits a high extent of disentanglement, which allows users to interactively perform a sequence of edits on an image. Our experiments on both GAN-generated and real images qualitatively demonstrate the high quality and advantages of our method. 
\end{abstract}
\section{Introduction}
\label{sec:introduction}

Generative Adversarial Networks, like StyleGAN \cite{StyleGAN1, StyleGAN2} and BigGAN \cite{BigGAN}, are capable of generating diverse high-resolution images, sometimes indistinguishable from real photos. In addition, substantial semantic meaning has been found in the latent space of trained GANs, which makes high-level semantic image editing possible. Semantic image editing using GANs \cite{GAN} has found a broad range of applications in Digital Art \cite{AestheticsOfNeuralNetworkArt}, Fashion \cite{FashionGarmentGAN}, Interior Design \cite{decorate_bedroom}, and Face Editing \cite{FaceApp}.

Several recent works control the semantics of the GAN-generated images by tweaking the latent code to perform global \cite{InterFaceGAN, GANSpace, UnsupervisedLatentSpaceGAN, ClosedFormFactorization, GANalyze, GANSteerability} or localized \cite{decorate_bedroom, EditingInStyleEDO, StyleSpace, lowrankGAN} image editing. Although a lot of progress has been made in global image editing, it remains challenging to disentangle the semantic attributes and thus control the local semantics of the image. Therefore, in this paper, we specifically focus on localized semantic image editing, where the goal is to control one semantic attribute of the image without changing other image parts.

State-of-the-art methods for finding semantically localized latent space directions rely on the first-order Taylor expansion of the generator network \cite{StyleSpace, lowrankGAN}, and thus assume a linear relation between the latent code and the generated image. Since this assumption is only valid in a close proximity to the original latent code, these methods \cite{StyleSpace, lowrankGAN} are limited in the range of local editing they can achieve. Meanwhile, ~\cite{StyleSpace} proposes a method to achieve disentanglement by exhaustively searching the latent space of StyleGAN~\cite{StyleGAN1, StyleGAN2}, and consequently cannot be applied to other GAN architectures. However, we are specifically interested in designing a framework which is not only agnostic to the GAN architecture, but also is able to effectively disentangle the semantic attributes. To this end, we propose \textit{Locally Effective Latent Space Directions} (LELSD), a framework to find latent space directions that affect local regions of the output image. We introduce a novel objective function to evaluate the localization and disentanglement of an image edit by incorporating supervision from a pre-trained semantic segmentation model. Note that, the supervision could also come from unsupervised \cite{EditingInStyleEDO} or weakly-supervised \cite{LinearSemSegGAN} models that use the intermediate featuremaps of the generator network to achieve semantic segmentation. As a result, our method is not limited to any specific dataset. Figure~\ref{fig:main} shows some of the semantic edits that our method can perform. Since we apply optimization instead of exhaustive search, our training time is three orders of magnitude faster than \cite{StyleSpace}. Meanwhile, unlike \cite{lowrankGAN} we do not perform test-time optimization and thus allow interactive image editing.

\vspace{-5pt}
\section{Related Works}
\label{sec:related_works}
\newcommand\RotText[1]{\rotatebox{90}{\parbox{2cm}{\centering#1}}}

\begin{table}[t]
\resizebox{\linewidth}{!}{%
\begin{tabular}{c|ccccccccl}
\begin{tabular}[c]{@{}c@{}}\textbf{GAN-based Image} \\ \textbf{Editing Algorithms}\end{tabular} & 
\begin{tabular}[c]{@{}c@{}}\textbf{Supervision}\end{tabular} &
\begin{tabular}[c]{@{}c@{}}{\textbf{A}}\end{tabular} &
\begin{tabular}[c]{@{}c@{}}{\textbf{B}}\end{tabular} 
&\begin{tabular}[c]{@{}c@{}}{\textbf{C}}\end{tabular} & \begin{tabular}[c]{@{}c@{}}{\textbf{D}}\end{tabular} & \begin{tabular}[c]{@{}c@{}}{\textbf{E}}\end{tabular} & \begin{tabular}[c]{@{}c@{}}{\textbf{F}}\end{tabular} & 
\begin{tabular}[c]{@{}c@{}}{\textbf{G}}\end{tabular} & \\
\hline \hline
\cite{GANSpace, UnsupervisedLatentSpaceGAN, ClosedFormFactorization}                                                                      & Unsupervised    & \cmark                                                            & \cmark & \cmark                                                                                                  & \cmark                                                                 & \cmark                                                                   & \xmark                                                                  & \xmark & \multirow{8}{*}{\rotatebox{90}{\parbox{3cm}{\centering \textbf{\hspace{3pt} Latent Space Traversal}}}} \\

Zhu et al. \cite{lowrankGAN}                                                                     & Unsupervised      & \cmark                                                   & \xmark                                                               & \cmark                                                            & \cmark                                                                 & \xmark                                                                   & \cmark                                                                  & \xmark  &                                                                 \\
\cite{ContinuousFactorVariationGAN, GANSteerability}                                              & Self-Supervised & \cmark                                                            & \cmark & \cmark                                                                                                             & \cmark                                                                 & \cmark                                                                   & \xmark                                                                  & \xmark &                                                                   \\
InterFaceGAN \cite{InterFaceGAN}                                                                 & Supervised      & \xmark                                                             & \cmark  & \cmark                                                                                                           & \cmark                                                                 & \cmark                                                                  & \xmark                                                                  & \xmark &                                                                  \\

GANALYZE \cite{GANalyze}                                                                     & Supervised      & \cmark                                                            & \cmark & \cmark                                                                                               & \cmark                                                                 & \cmark                                                                   & \xmark                                                                  & \xmark  &                                                                 \\
StyleSpace \cite{StyleSpace}                                                                     & Supervised      & \cmark                                                            & \cmark                                                             & \xmark                                                            & \cmark                                                                 & \xmark                                                                   & \cmark                                                                  & \cmark  &                                                                 \\
\textbf{LELSD (Ours)}                                                                   & Supervised      & \cmark                                                 & \cmark                                                              & \cmark                                                            & \cmark                                                                 & \xmark                                                                  & \cmark                                                                 & \cmark &            \multirow{8}{*}{\rotatebox{90}{\parbox{3cm}{\centering \textbf{\hspace{0pt} Image Composition}}}} \\
\hline \hline
Bau et al. \cite{RewritingGAN}                                                      & Unsupervised    & \cmark                                                      & \xmark                                                              & \xmark                                                            & \xmark                                                                  & \xmark                                                                  & \cmark                                                                 & \cmark &  \\
Chai et al. \cite{gan_compositionality}                                                      & Unsupervised    & \cmark                                            & \cmark                                                              & \cmark                                                            & \xmark                                                                  & \xmark                                                                  & \cmark                                                                 & \cmark &                                                                   \\
Editing in Style \cite{EditingInStyleEDO}                                                             & Unsupervised    & \cmark                                             & \cmark                                                               & \xmark                                                             & \xmark                                                                  & \xmark                                                                   & \cmark                                                                 & \xmark  &                                                                  \\
Zhang et al. \cite{decorate_bedroom}                                                  & Supervised      & \xmark                                          & \cmark                                                             & \xmark                                                             & \xmark                                                                  & \xmark                                                                  & \cmark                                                                 & \cmark  & \\

Barbershop \cite{BarbershopGI}       & Supervised    & \cmark                                                  & \cmark                                                              & \xmark                                                             & \xmark                                                                  & \xmark                                                                  & \cmark                                                                 & \xmark &  
\end{tabular}
}
\vspace{-5pt}
\caption{Comparison of GAN-based image editing algorithms by their characteristics.  \textbf{(A)} Works on any Dataset,  \textbf{(B)} Does not need test-time optimization,  \textbf{(C)} Works on any GAN architecture, \textbf{(D)} Can perform the edit using a single image,  \textbf{(E)} Allows global semantic editing,  \textbf{(F)} Allows localized semantic editing,  \textbf{(G)} Allows editing any object in the image.\vspace{-10pt}} 
\label{image_edit_comparison}
\end{table}

Table~\ref{fig:main} summarizes the strengths and weaknesses of of different GAN-based image editing methods. Some works use an unsupervised approach to discover meaningful latent space directions, and then manually attribute a semantic meaning to each of the found directions \cite{GANSpace, UnsupervisedLatentSpaceGAN, ClosedFormFactorization, ContinuousFactorVariationGAN}. However, the discovered directions are semantically entangled and usually change more than one attribute simultaneously. Hence, they are not suitable for localized image editing.

To solve this problem, \cite{InterFaceGAN, GANalyze, GANSteerability} use an external supervision and find latent space directions that yield the desired change in the generated images. This is done by finding the latent space direction that maximizes a designed objective function.

There are two distinct approaches to GAN-based localized semantic image editing: 1) \textit{Latent Space Traversal}, and 2) \textit{Image Composition}. In the former, the goal is to discover latent space directions that yield localized changes in the output image. The later aims to combine different parts from two images to achieve localized editing, e.g., transferring the nose from one face image to another. Our method falls in the first category. The disadvantage of the Image Composition methods is that they require a second image to transfer the parts from (See Table~\ref{image_edit_comparison}, Column D). In this paper we thus focus on the Latent Space Traversal methods as they can perform single-image editing, which is more user-friendly.

InterFaceGAN \cite{InterFaceGAN} finds latent space directions that maximally change the score of a pre-trained SVM classifier for face attributes, and therefore is only applicable on face images. On the other hand, \cite{StyleSpace, lowrankGAN} use the gradient of the output image w.r.t the latent code to find subspaces of the latent code that highly correlate with local regions in the generated image. Wu et al. \cite{StyleSpace} use a pretrained semantic segmentation network to find channels in StyleGAN's style code that have a high overlap with a semantic category, e.g., eyes. Zhu et al \cite{lowrankGAN} perform test-time optimization to find latent space directions that mostly affect the regions of the image outline in the user's query. The upper part of Table~\ref{image_edit_comparison} compares the existing  Latent Space Traversal methods in the literature. 

Similar to \cite{decorate_bedroom, StyleSpace, BarbershopGI}, we use the supervision from a pretrained semantic segmentation network, and propose a novel scoring function that encourages the latent space directions that mainly affect the desired semantic part to edit, e.g., eyes in a face image. Meanwhile, we allow both coarse and fine-grained semantic changes by adopting the layer-wise editing approach of \cite{GANSpace} for both style-based GANs and BigGAN. GAN Inversion is complementary to our method and advancement in GAN Inversion research \cite{Image2StyleGAN++, Richardson2020EncodingIS} also enhances the quality of semantic editing of real photos when combined with our method. 

\vspace{-5pt}
\section{Method}
\label{sec:method}
\begin{figure*}[t]
\begin{center}
  \includegraphics[trim={60pt 105pt 130pt 109pt},clip,width=\textwidth]{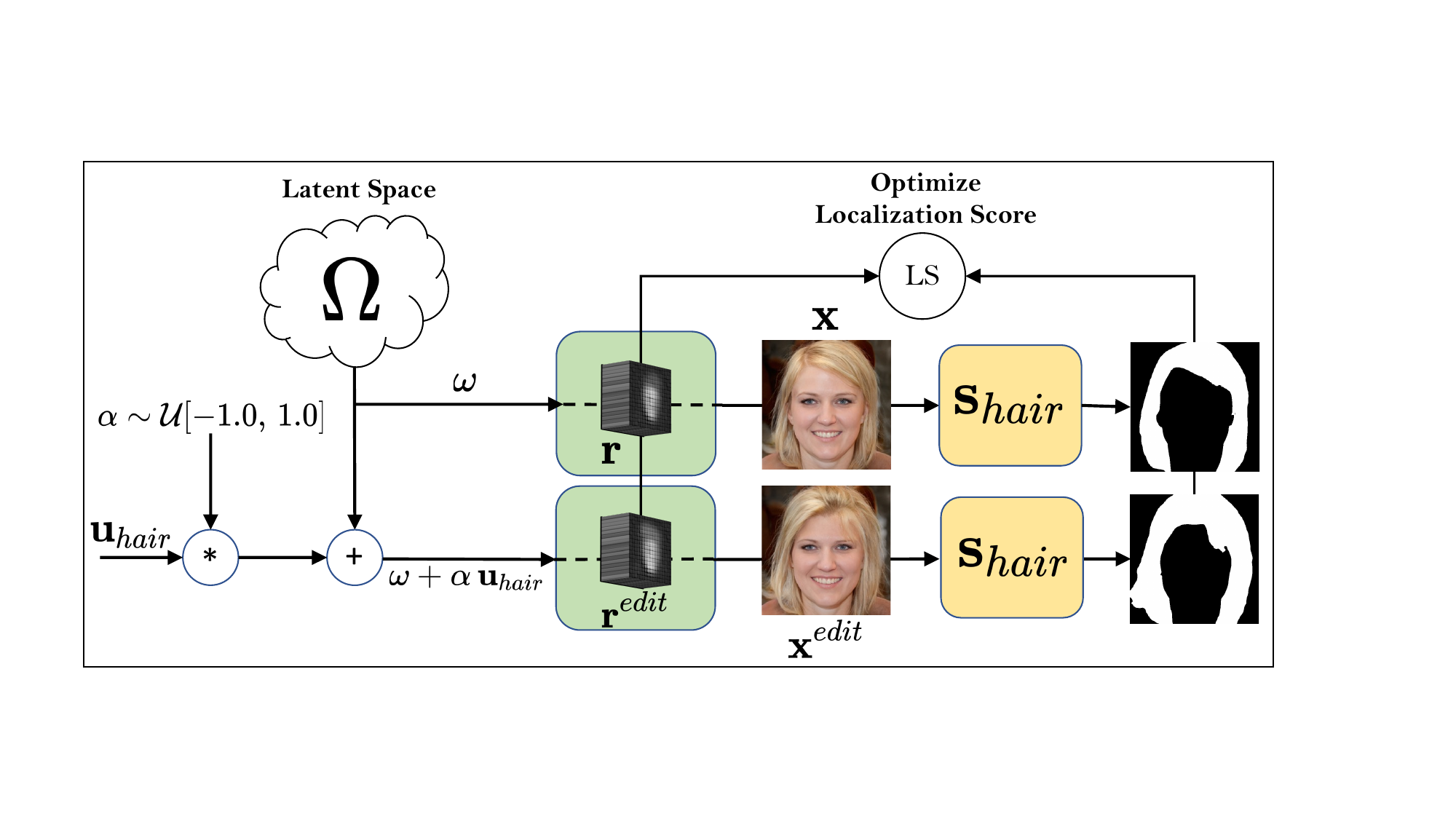}
\end{center}
    \vspace{-15pt}
  \caption[skip=0pt]{Scheme of our method. The green boxes indicate the pretrained generator network and the yellow box shows the pretrained semantic segmentation model. We draw random samples from the latent space $\Omega$ and optimize the latent space direction to maximize the localization score. }
    \label{fig:scheme_of_method} 
\vspace{-8pt}
\end{figure*}

Figure~\ref{fig:scheme_of_method} provides an overview of our method. The generator network $G(.)$ in a GAN generates an image starting from a latent code $\mathbf{\omega} \in \Omega$, i.e. $\mathbf{x} = G(\boldsymbol{\omega}) = f(h(\boldsymbol{\omega}))$ where $\mathbf{r} = h(\boldsymbol{\omega})$ is a tensor representing the activation of an intermediate layer in the network. The latent space    $\Omega$ can be any of $\mathcal{Z}, \mathcal{W}, \mathcal{W}+, \mathcal{S}$ for the StyleGAN generator as in \cite{StyleSpace}, and $\mathcal{Z}, \mathcal{Z}+$ for BigGAN generator as in \cite{GANSpace}. Semantic editing of an image is done by moving its latent code along a specific direction 
\begin{equation}
        \label{eqn:gan_edit}
        \mathbf{x}^{edit}(\mathbf{u}) = f(\mathbf{r}^{edit}(\mathbf{u})) = G(\boldsymbol{\omega} + \alpha \mathbf{u})
\end{equation}
where $\alpha$ controls the intensity of the change, and the latent direction $\mathbf{u}$ determines the semantic of the edit. 

Our goal is to find an editing direction $\mathbf{u}_c$ that mostly changes parts of the generated image corresponding to a binary mask given by a pretrained semantic segmentation model $\mathbf{s}_c(\mathbf{x})$ where $c$ indicates the desired object to edit in the image. Based on this, we can write the localization score as
\begin{equation}
        \label{eqn:localization_score}
        LS(\mathbf{u}) = \frac{\sum_{i,j} \overset{\downarrow}{\tilde{\mathbf{s}_c}}  (\mathbf{x}, \mathbf{x}^{edit}) \odot |\mathbf{r} - \mathbf{r}^{edit}(\mathbf{u})|^2}{\sum_{i,j} |\mathbf{r} - \mathbf{r}^{edit}(\mathbf{u})|^2} 
\end{equation} where $i, j$ iterate over the spatial dimensions and $\overset{\downarrow}{\tilde{\mathbf{s}_c}}  (\mathbf{x}, \mathbf{x}^{edit})$ is the average of the two semantic segmentation masks down-sampled to the resolution of the corresponding layer. 
This objective function measures the proportion of the change in the featuremap that happens inside the semantic segmentation mask. Our final objective function is calculated by simply summing up the localization scores for all intermediate layers in the generator network. Unlike \cite{lowrankGAN} that only aims to achieve localized change in the generated image, we also encourage the intermediate featuremaps to only change locally. This allows us to achieve a larger variety of edits than \cite{lowrankGAN}. For example, we can change both hairstyle and hair color, while \cite{lowrankGAN} cannot manipulate hairstyle.

\vspace{-5pt}
\section{Experiments}
\label{sec:experiments}
For the pretrained semantic segmentation model, we use Face-BiSeNet \cite{BiSeNet} for face/portrait images and DeepLabV3 \cite{DeepLabV3} for other images. We use the Adam optimizer to find a latent space direction that maximizes the localization score defined in Equation~\eqref{eqn:localization_score}\footnote{Our code can be found at  https://github.com/IVRL/LELSD}. We train on 800 randomly sampled latent codes with a batch size of 4, starting from a learning rate of 0.001 and halving it every 50 steps. Optimizing a latent space direction for each semantic part takes approximately two minutes on a single Tesla V100 GPU. We observe that our method is robust to the choice of the mask aggregation method in Equation~\eqref{eqn:localization_score} and works as well with the union or the intersection of the two masks.

\subsection{Finding multiple directions}

\begin{figure}[bthp]
\begin{center}
  \includegraphics[trim={8pt 12pt 11cm 8pt},clip,width=1.05\linewidth]{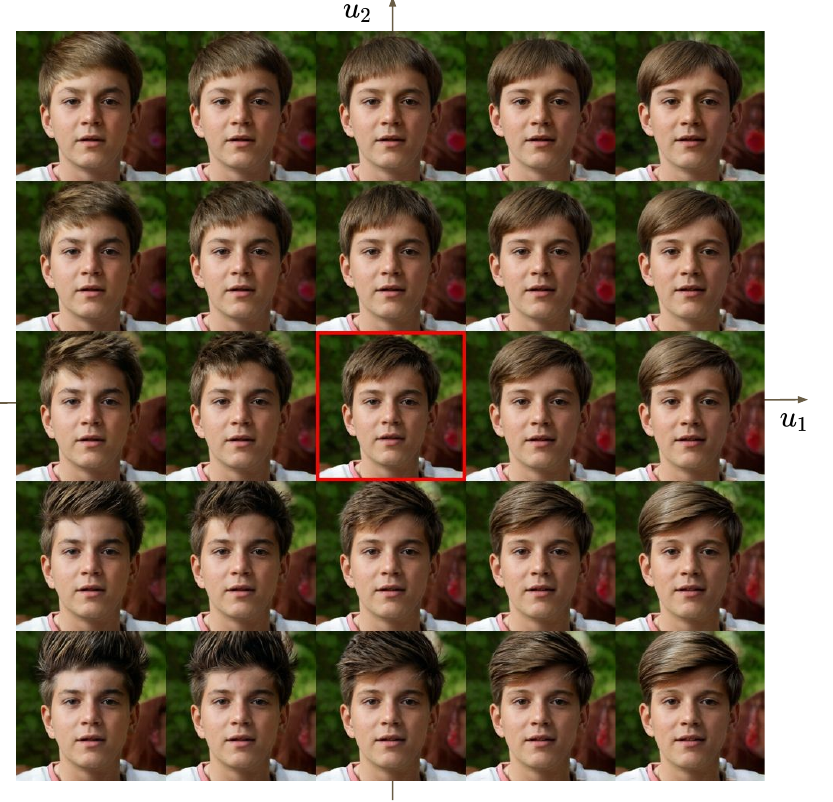}
\end{center}
    \vspace{-20pt}
  \caption[skip=0pt]{Visualization of two latent space directions found by our method for editing hairstyle. Linearly combining different directions gives limitless image editing possibilities to users.}
\label{fig:hairstyle}
\vspace{-8pt}
\end{figure}

In order to find two or more distinct directions for editing the same semantic part such as hair, we add $R(\mathbf{u}_1, \;..., \mathbf{u}_k) = \frac{-1}{2}||\textup{Corr}(\mathbf{u}_1, \;..., \mathbf{u}_K) - \textup{I}_K||_{\textup{F}}$ as a regularization term to our objective function, where $\textup{Corr}(.)$ is the correlation matrix of a set of vectors, $||.||_{\textup{F}}$ is the Frobenius Norm, and $\textup{I}_K$ is the $K\times K$ identity matrix. Our final objective can thus be written as 
\begin{equation}
        \label{eqn:final_objective}
        J(\mathbf{u}_1, \;..., \mathbf{u}_k) = \sum_{k} LS(\mathbf{u}_k) + cR(\mathbf{u}_1, \;..., \mathbf{u}_k)
\end{equation} where $c$ is the regularization coefficient. The added regularization term encourages the editing directions to be mutually perpendicular, and carry distinct semantics as can be seen in Figure~\ref{fig:hairstyle}. We linearly increase the number of training samples w.r.t  $K$ and alternate between each $\mathbf{u}_k$ during the optimization process. 

\subsection{Comparison with First-Order methods}
\begin{figure}[t]
\begin{center}
  \includegraphics[trim={40pt 0pt 0pt 0pt},clip,width=1.05\linewidth]{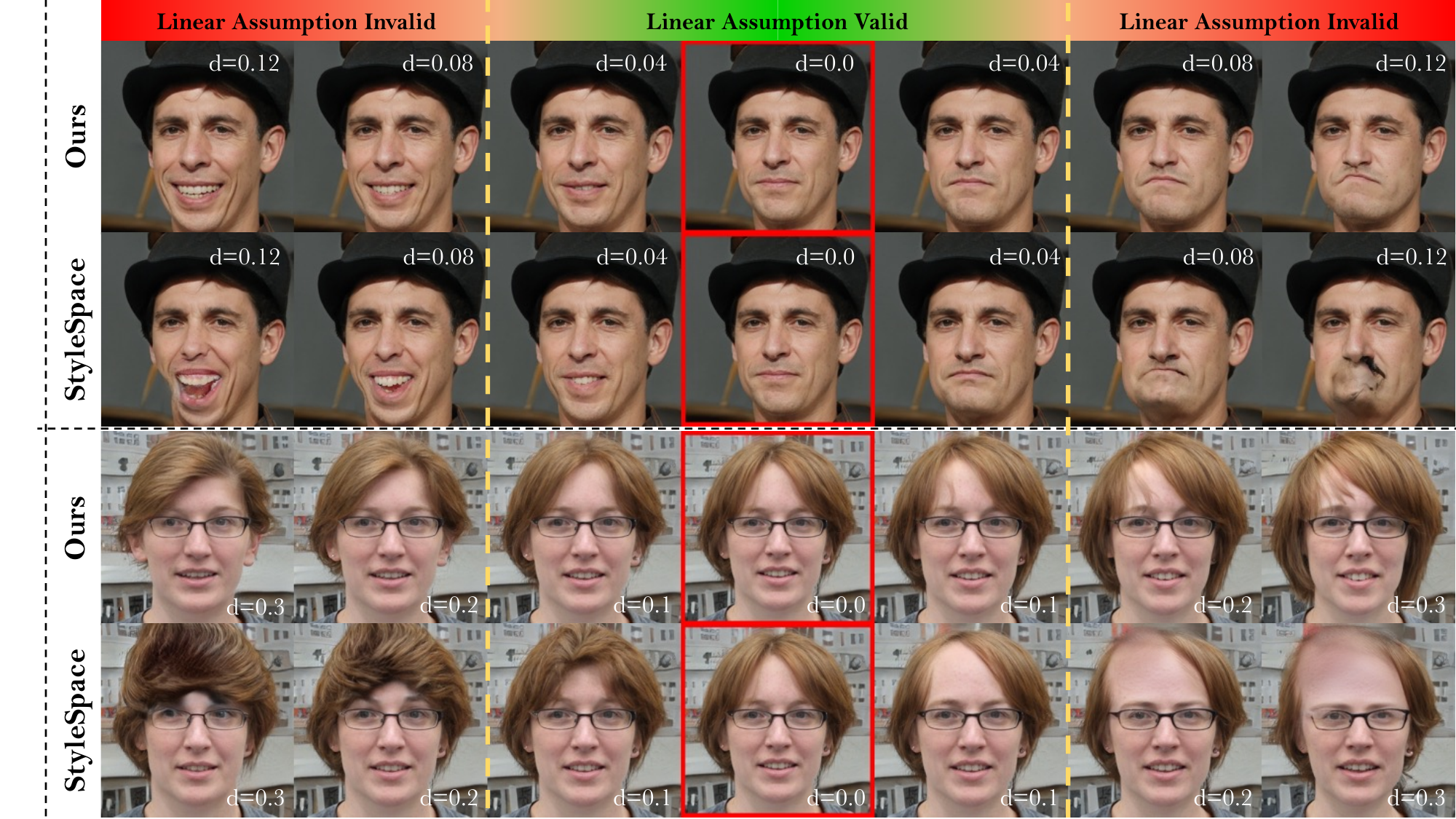}
\end{center}
    \vspace{-15pt}
  \caption[skip=0pt]{Comparison of our method with StyleSpace \cite{StyleSpace}. The first-order approximation of the generator network is only valid in a close proximity of the original latent code. Hence, gradient-based methods like StyleSpace \cite{StyleSpace} perform poorly as the editing strength increases. The $d$ in the figure shows the LPIPS distance between the edited and original images.}
\label{fig:comparison}
\vspace{-8pt}
\end{figure}

Both \cite{StyleSpace, lowrankGAN} rely on the first-order Taylor expansion of the generator network and assume a linear relationship between the generated image and the latent code. This causes them to perform poorly as the editing strength $\alpha$ increases.
Since $\alpha$ in Equation~\eqref{eqn:gan_edit} has a different scale for each GAN-based image editing method, we use the LPIPS distance \cite{LPIPS} for the comparison. For each method we find the value of $\alpha$ such that $LPIPS(\mathbf{x}, \mathbf{x}^{edit}) = d$, and show that for higher values of $d$ where the linearity assumption is not valid, our method outperforms \cite{StyleSpace}. Note that there are two values of $\alpha$ that yield the same LPIPS distance $d$, where one is positive and the other one is negative. Figure~\ref{fig:comparison} compares the edits performed on mouth and hair by our method and StyleSpace \cite{StyleSpace}, for different values of editing strength. As $d$ and subsequently absolute value of $\alpha$ increase, our method performs coherently while StyleSpace distorts the semantics of the image. 

\subsection{Editing Real Images}
\begin{figure}[!t]
\begin{center}
  \includegraphics[trim={0pt 0pt 94pt 0pt},clip,width=1.05\linewidth]{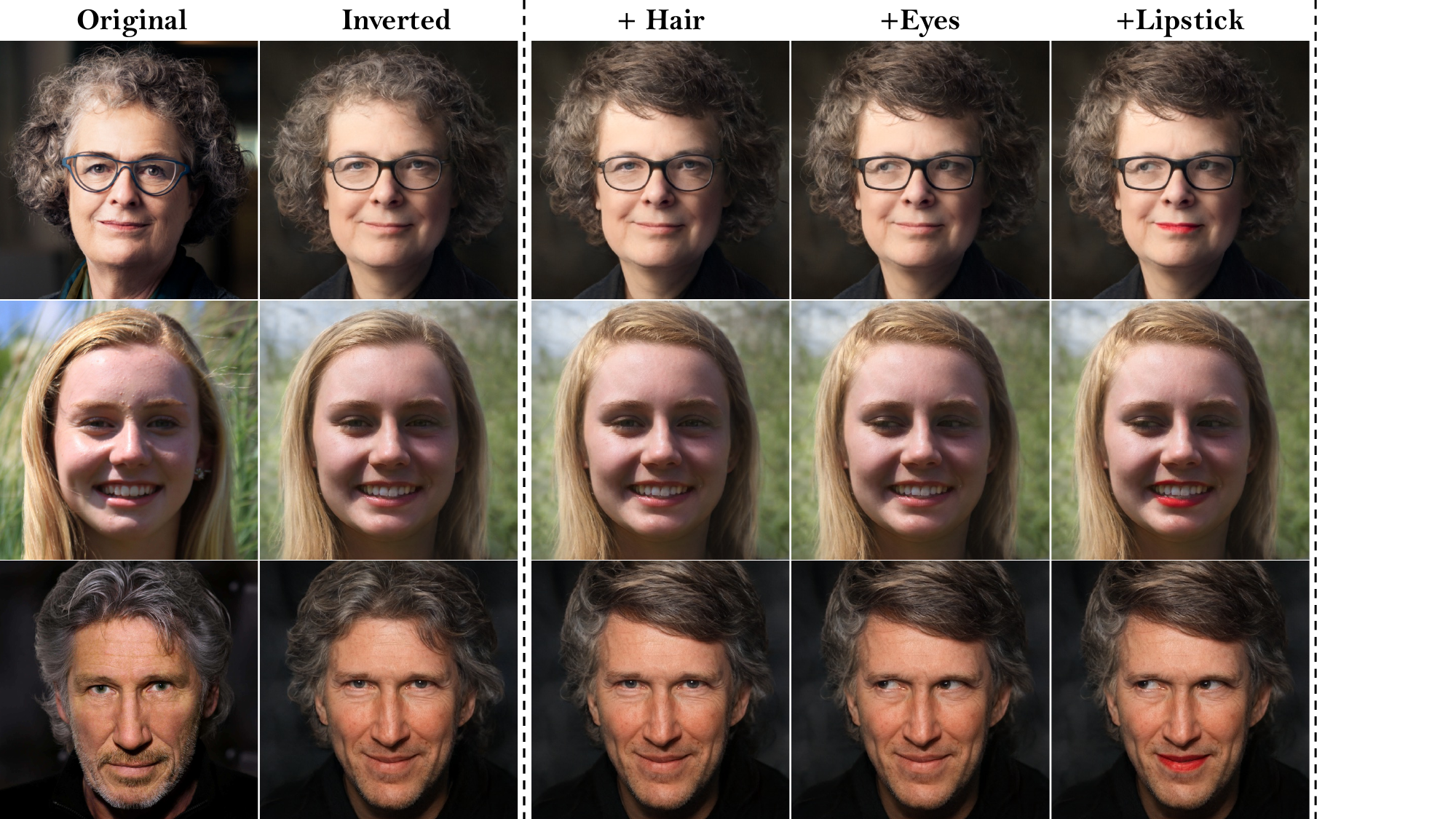}
\end{center}
    \vspace{-15pt}
  \caption[skip=0pt]{Sequential editing applied to real images. The edits are semantically consistent between different images and each edit only changes the desired part without affecting previous changes.}
\label{fig:sequential_editing}
\vspace{-8pt}
\end{figure}

Combined with a GAN Inversion model, our method allows editing real images. We use e4e \cite{Richardson2020EncodingIS} trained on StyleGAN2 FFHQ to project real face photos into the latent space of the StyleGAN.
More importantly, we can perform sequential editing by simply adding up the discovered latent space directions for each semantic. Figure~\ref{fig:sequential_editing} shows a series of edits applied to the inversion of real photos. As can be seen, the semantics of the edits are consistent across different images. The quality of the GAN inversion is beyond the scope of this paper.

\subsection{Performance Comparison}
Interactiveness and performance are two very important factors in the image editing experience. 
The method in \cite{lowrankGAN} requires test-time optimization and hence is not interactive. Although the approach of \cite{StyleSpace} allows interactive editing, it requires a lot of training time as it needs to separately backpropagate through all 6080 channels in the style code of the StyleGAN. As we use the same number of training samples as \cite{StyleSpace}, we estimate that our method is three orders of magnitude faster than \cite{StyleSpace}.

\vspace{-5pt}
\section{Conclusion}
\label{sec:conclusion}
In this work, we have presented LELSD, our computationally friendly framework that uses the supervision from a pre-trained semantic segmentation network to maximize a novel objective function that encourages local image edits and can be applied to any GAN architecture and dataset. Our experiments in different setting qualitatively show the advantage of our method, especially in the extent of disentanglement achieved between local attributes.

\bibliographystyle{IEEEbib}
\bibliography{main}

\end{document}